\documentclass[letterpaper, 10 pt, conference]{ieeeconf}  

\IEEEoverridecommandlockouts                              

\usepackage{amsmath} 
\usepackage{amssymb}  
\usepackage{mathrsfs}
\usepackage{graphbox}
\usepackage{subcaption}
\usepackage{float}
\usepackage{multicol, blindtext}
\usepackage{color}
\usepackage{soul}
\usepackage{graphicx}  
\usepackage{balance}
\usepackage{dsfont}
\usepackage{hyperref}
\title{\LARGE \bf
Learning Socially Appropriate Robot Approaching Behavior Toward Groups using Deep Reinforcement Learning
}

\author{Yuan Gao$^{1}$, Fangkai Yang$^{2\dagger}$, Martin Frisk$^{1\dagger}$, Daniel Hernandez$^{3}$, Christopher Peters$^{2}$ and Ginevra Castellano$^{1}$ 
\thanks{$^{\dagger}$ Authors contributed equally.}
\thanks{*This work is partly supported by the Swedish Research Council (grant n. 2015-04378) and partly by the COIN project (RIT15-0133) funded by the Swedish Foundation for Strategic Research.}
\thanks{$^{1}$ Author is with Department of Information Technology,
        Uppsala University, Uppsala, Sweden.
        {\tt\small alex.yuan.gao@it.uu.se}}
\thanks{$^{2}$ Author is with Department of Computational Science and Technology,
        KTH Royal Institute of Technology, Stockholm, Sweden.}
\thanks{$^{3}$ Author is with Department of Computer Science,
        University of York, York, United Kingdom.}    
}%

\begin{document}

\maketitle
\thispagestyle{empty}
\pagestyle{empty}
\begin{abstract}

Deep reinforcement learning has recently been widely applied in robotics to study tasks such as locomotion and grasping, but its application to social human-robot interaction (HRI) remains a challenge. In this paper, we present a deep learning scheme that acquires a prior model of robot approaching behavior in simulation and applies it to real-world interaction with a physical robot approaching groups of humans. The scheme, which we refer to as Staged Social Behavior Learning (SSBL), considers different stages of learning in social scenarios. We learn robot approaching behaviors towards small groups in simulation and evaluate the performance of the model using objective and subjective measures in a perceptual study and a HRI user study with human participants. Results show that our model generates more socially appropriate behavior compared to a state-of-the-art model. 


\end{abstract}
\section{Introduction}
Deep reinforcement learning (DRL) algorithms provide a framework for automatic robot perception and control~\cite{sunderhauf2018limits}~\cite{chernova2014robot}. In recent years, methods based on DRL have achieved great performance in different control tasks such as grasping and locomotion~\cite{levine2016end}. However, the question of how to make robots learn appropriate social behaviors under modern frameworks remains underexplored, partly due to the lack of cross-disciplinary synergies in human-robot interaction (HRI) studies. As a consequence, the interaction scenarios studied in previous research have been limited to simplified cases and the algorithms studied to relatively simple ones~\cite{ferreira2015reinforcement}. 

A promising, but underexplored approach, to robot learning in social HRI scenarios is to learn a prior model in simulation first and then refine the learned policy using model-based reinforcement learning (RL) in the real-world. Learning a prior model in a simulated environment has a lot of potential benefits. First, it can save a significant amount of real-world interactions. Several works~\cite{hanna2017grounded}~\cite{bousmalis2017using} have shown that learning a model for physical interactions can help robots learn faster in the real-world. Secondly, in social interactions, humans have little tolerance for random behaviors~\cite{abubshait2017you}, and lose interest quickly if the model deviates too much from social norms. Additionally, the mathematical modeling of social interactions in a simulated setting allows researchers to control factors more rigorously, which can help with the issue of replicability. 
However, unlike simulating physical interactions~\cite{jonschkowski2014state}, simulating social HRI poses a different set of challenges.  One of the main challenges is that it is hard for the simulator to accurately model human relevant behavior. Simulators of physical interactions are based on physical laws which are well understood, while human behavior is less predictable. Nevertheless, two ways have been considered to simulate social feedback based on real world signals. The first one is to use computational models~\cite{jan2007dynamic} that have been studied in experiments~\cite{pedica2010avatars} and the second one is to use machine learning methods~\cite{yuan2018human}. 



\begin{figure}[t] 
\centering
\includegraphics[align=c,width=.98\linewidth]{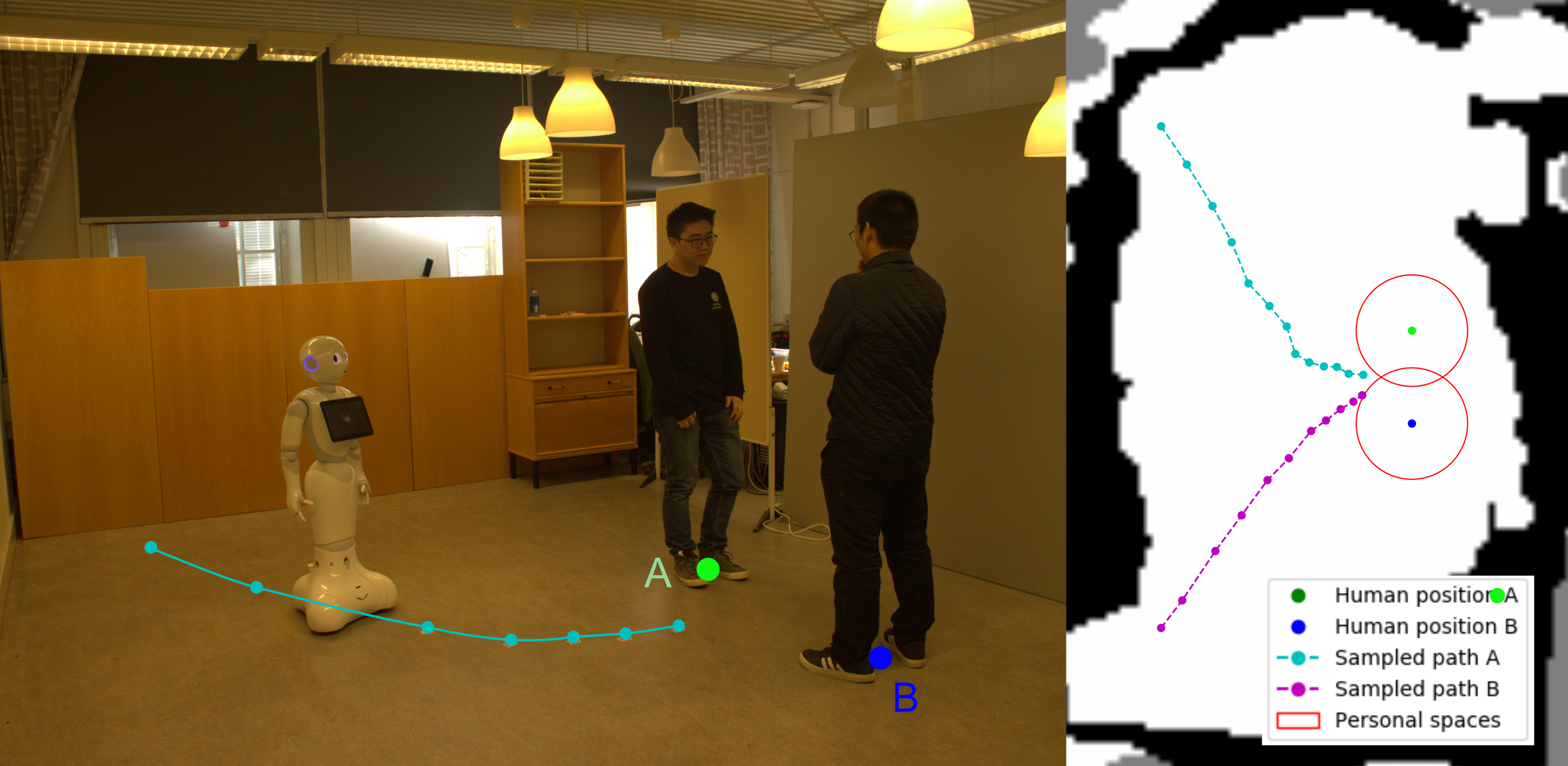}

\caption{Pepper robot approaching a group of people (left); top view of robot's paths generated using our proposed SSBL deep learning scheme, projected onto a map constructed using simultaneous localization and mapping (right). 
}\label{physical_paths}
\end{figure}

In this paper, we propose a deep learning scheme, called Staged Social Behavior Learning (SSBL), for learning robot appropriate social behavior with continuous actions in a simulated environment, and we apply it to real-world interaction with a physical Pepper\footnote{https://www.softbankrobotics.com/emea/en/pepper} robot interacting with humans. Specifically, we consider a task in which the robot moves toward a small group, positioned in an F-formation~\cite{kendon1990conducting}, based on its simulated social feedback, using the Social Force Field Model (SFFM) \cite{pedica2008social} . The task is learned in an end-to-end fashion, i.e., from vision to social behaviors in a virtual environment. SSBL involves a pipeline for simulated social robot learning that deconstructs a social task into three steps. In the first step, the robot learns a compressed representation of the world from vision or other modalities. This step is important because it significantly reduces the complexity of the DRL problem. After the compressed information is learned, the algorithm learns a dynamical model from a prior model  which is built upon social forces \cite{jan2007dynamic} \cite{pedica2008social} in the environment. The last step is to make sure that the learned behavior follows the social standard by using simulated social norms as realistic reward.
{In this study,} {we focus on the first two steps of the SSBL framework and }learn robot approaching behaviors towards small groups in simulation and evaluate the performance of the model using objective and subjective measures in a perceptual study and a HRI user study with human participants. Figure~\ref{physical_paths} shows our physical robot experiment setup. The code of this project is publicly available on the GitHub.\footnote{The code is publicly available at \href{https://github.com/gaoyuankidult/PepperSocial/tree/master}{https://github.com/gaoyuankidult/Pepp
erSocial/tree/master}}



\section{Literature Review} 

\subsection{Deep Reinforcement Learning in HRI}
Reinforcement learning (RL) has been used since the early days of HRI. One of the first works that considered using social feedback as accumulative rewards was conducted by Bozinovski~\cite{bozinovski1996emotion}~\cite{bozinovski1982learning}. After that, many papers in HRI started to investigate the effect of RL algorithms such as Exp3~\cite{yuan2018when} or Q-learning~\cite{mataric1997learning} in social robotics settings. However, since such algorithms lack the ability to capture important features from high-dimensional signals~\cite{mnih2015human}, their applicability to solve HRI problems remains limited. After the era of deep learning started in 2006~\cite{lecun2015deep}, many different algorithms were proposed to understand different modalities in HRI, for example, ResNet~\cite{he2016deep} for image processing and Long Short-Term Memory~\cite{hochreiter1997long} (LSTM)-based solutions for text processing. As a consequence, some HRI researchers started to investigate deep learning's role in the area of HRI. A pioneering work was conducted by Qureshi in 2017~\cite{qureshi2016robot}. In this work, a Deep Q-Network (DQN)~\cite{mnih2015human} was used to learn a mapping from visual input to one of several predefined actions for greeting people. Another work was conducted by Madson~\cite{clark2018deep}, where a DQN was used for learning generalized, high-level representations from both visual and auditory signals. 

\subsection{Learning Representations}
RL based solely on visual observations has been used to solve complex tasks such as playing ATARI games~\cite{mnih2013playing}, driving simulated cars~\cite{tan2018autonomous} and navigating mazes~\cite{mirowski2016learning}. However, learning policies directly from high-dimentional data such as images requires a large amount of samples, which makes it intractable in social robot learning~\cite{bohmer2015autonomous}. One solution is to use low-dimensional hand-crafted features as the state, but this would reduce learning autonomy. 

Prior works have utilized deep autoencoders (AEs) to learn a state representation, including Lange et al.~\cite{lange2012autonomous}. Several variants of AEs have been applied as well, including attempts by B{\"o}hmer et al.~\cite{bohmer2015autonomous} to learn the dynamics of the environment by constructing an AE predicting the next image, and Finn et al.~\cite{finn2015deep} who adopted a spatial AE (SAE) to learn an intermediate representation consisting of image coordinates of relevant features. The latter suggested that this intermediate representation made it particular well suited for high-dimensional continuous control. 

\subsection{Modelling Groups and Robot Approaching Behavior}
Numerous works have been done in group dynamic behaviors. Particle-based methods \cite{heigeas2010physically} \cite{treuille2006continuum} simulate global collective behaviors of large scale groups or crowds. For modeling small scale groups, agent-based methods \cite{musse2001hierarchical} \cite{reynolds1999steering} are adopted to simulate the behavior of each individual based on rules of behavior. Specifically, in a small multi-party conversation group, Kendon~\cite{kendon1990conducting} proposed the \textit{F-formation} system to define the positions and orientations of individuals within a group, which characterized dynamic group behaviors. Several studies have been carried out that concern robot approaching behaviors towards small groups i.e. in which an agent moves towards a group in an attempt to join an ongoing task or conversation. Ram{\'\i}rez et al.~\cite{ramirez2016robots} adopted inverse reinforcement learning, involving several participants demonstrating approaching behaviors for a robot to learn. Pedica et al.~\cite{pedica2012lifelike} integrated behavior trees in their reactive method to simulate lifelike social behaviors, including robot approaching behavior towards groups. Both approaching and leaving behaviors are considered in~\cite{yang2017expressive}, where a finite state machine is utilized in the transitions between different social behaviors. Jan et al.~\cite{jan2007dynamic} presented an algorithm for simulating movement of agents, such as an agent joining the conversation. The agents dynamically move to new locations, but without proper orientations. More recently, Samarakoon et al. \cite{samarakoon2018replicating} designed a method to replicate the natural approaching behaviors of humans. Meanwhile, a fuzzy inference system was proposed in \cite{bhagya2018proxemics} to decide the approaching proxemics based on the behaviors of the user.


\begin{figure}[b] 
\centering
\includegraphics[width=.46\linewidth]{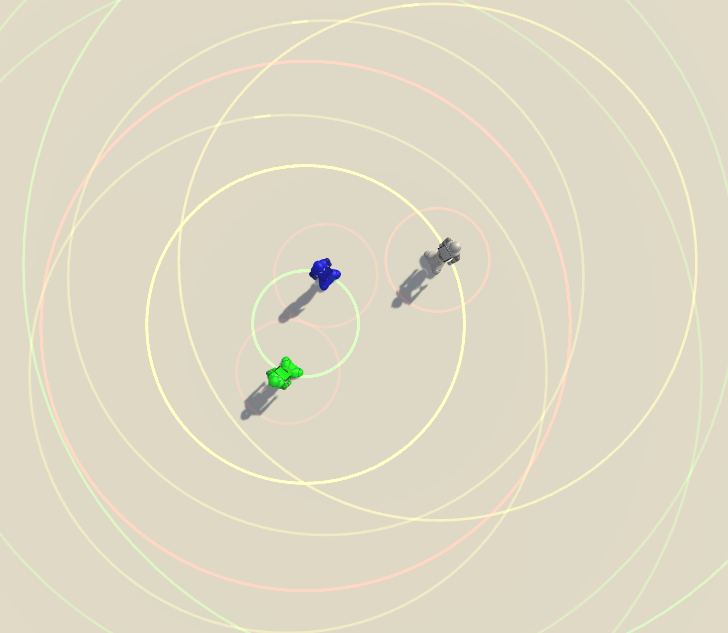}
\includegraphics[width=.40\linewidth]{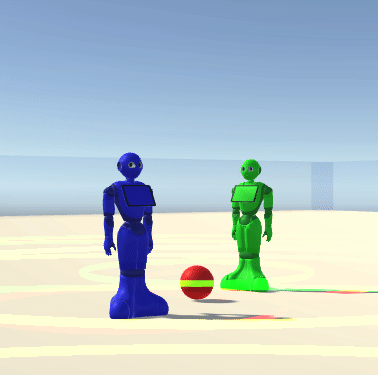}
\caption{Two views of the environment setup. The left image shows a top-down view of the environment.  The right image shows the first-person view from the robot's perspective. The environment contains two Simulated Human Agents (SHAs: green and blue), and a robot agent (gray). Circles around the human agents and the robot represent different levels of space as discussed in Section~\ref{subsection:approaching}.}
\label{views}
\end{figure} 

\section{Methodology}
In the following sections, we introduce the fundamental concepts needed to train a prior model for robot approaching behaviors in accordance with SSBL. Section \ref{section_preliminaries} introduces some basic concepts and details on how the environment was set up. Sections \ref{section_staterepresentation} to \ref{subsection:socialnorms} describe the three stages of SSBL training. Specifically Section \ref{section_staterepresentation} pertains to the state representation and its training procedure. It also details the various architectures used to evaluate this step. Section \ref{subsection:approaching} shows how one can formulate the training of a dynamical model within an RL framework. Section \ref{subsection:socialnorms} shows how social norms can be acquired by utilizing concepts from the SFFM~\cite{pedica2008social}.

In the original SFFM work~\cite{pedica2008social}, SFFM was also used to generate social agent behaviors, which will be used as a baseline for evaluating our learned policy.

%

\begin{figure*}[h]
\includegraphics[width=1\textwidth]{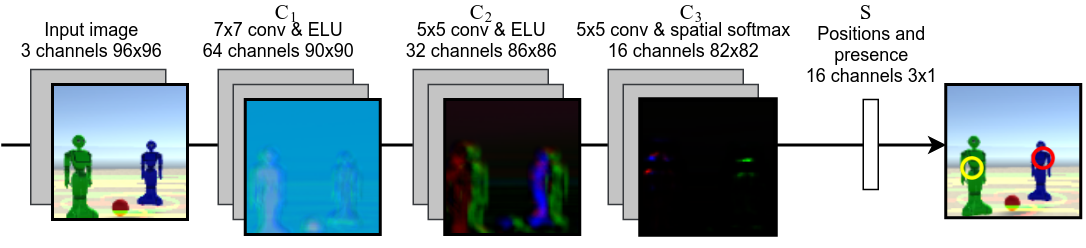}
\caption{Schematic view of the encoder used both by the architecture proposed by Finn et al.~\cite{finn2015deep} and ours. The rightmost image is not a part of the network, but a visualization of two of the positions computed in $S$. The slight discrepancy in position between the position of the activation peaks in $C_3$ and the circles in the output-image  is due to the convolutions not using any padding. This makes it so that $C_3$ only represents the center $82\times 82$ pixels of the input.}\label{ae_architecture}
\end{figure*}
\subsection{Environment Setup} \label{section_preliminaries}
In order to simulate robot approaching behaviors, we first build a simulator using \textit{Unity 3D}\footnote{https://unity3d.com/} game engine. The environment consists of a square floor surrounded by four walls. A conversation group which contains two Simulated Human Agents (SHAs) is spawned at a random position within this domain. The robot agent is spawned outside the group and performs approaching behaviors. The virtual agents (agents in the group and robot agent) are pre-defined assets which resemble the SoftBank Pepper robot. 
Figure~\ref{views} shows one example of environment's top-down and first-person view. The blue and green agents are SHAs and the gray one at the top right of the top-down view is the robot agent. The first-person view (Figure~\ref{views}, right) is from the robot agent's perspective.

In this paper, we are mainly concerned with the task of learning a prior model in the simulator for a robot's approaching behaviors towards small groups of individuals. The task can be formulated as an RL problem. Let us consider $\mathbf{s}_t$ and $\mathbf{a}_t$ as the state and action of the robot agent at time $t$, respectively. Learning the dynamic behavior for approaching a group can be viewed as maximizing the expected cumulative reward $E_{\tau\sim\pi}[\mathcal{R}(\tau)]$ over trajectories $\tau = \{\mathbf{s}_1,\mathbf{a}_1,\dots, \mathbf{s}_T, \mathbf{a}_T\}$, where $\mathcal{R}(\tau)=\sum_{t=1}^T\mathcal{R}(\mathbf{s}_t, \mathbf{a}_t)$ is the cumulative reward over $\tau$. The expectation is under distribution $p(\tau) = p(\mathbf{s}_1)\prod_{t=1}^{T}p(\mathbf{s}_{t+1}|\mathbf{s}_t,\mathbf{a}_t)p(\mathbf{a}_t|\mathbf{s}_t)$, where $\pi(\mathbf{s}_t) = p(\mathbf{a}_t|\mathbf{s}_t)$ is the policy we would like to train and $p(\mathbf{s}_{t+1}|\mathbf{s}_t,\mathbf{a}_t)$ is the forward model determined by the environment.


\subsection{State Representations} \label{section_staterepresentation}
In our experiments, we try three modes of representing the environment state to the robot agent: \emph{Vector}, \emph{CameraOnly} and \emph{CameraSpeed}. The first mode is a vector-based representation, consisting of the positions and velocities of all the agents, together with the positions of the walls. This representation is ideal for learning, so it serves as an upper bound on the performance of this task. 

The second and third modes are designed to resemble two common robotic settings: one where the robot is equipped with a camera, and one where the robot has both a camera and the ability to estimate its speed. In these modes, the full states are given as $\mathbf{s}_t = \mathbf{I}_t$ and $\mathbf{s}_t = (\mathbf{I}_t, \mathbf{v}_t)$ respectively. Here $\mathbf{I}_t$ is the visual information from the robot's first-person view  rendered by the Unity engine, and $\mathbf{v}_t$ the velocity of the robot.





The method employed in this work is to learn a mapping from input images to simplified low-dimensional state representations, thus circumventing some of the problems associated with RL from high dimensional input~\cite{mnih2013playing}. To do this, we utilize an autoencoder (AE)~\cite{goodfellow2016deep}, a neural net $\phi$ that maps inputs to itself, s.t. $\phi(x) \approx x$. An AE can be decomposed into an encoder and a decoder, $\phi \equiv \phi_{dec} \circ \phi_{enc}$. By choosing the intermediate representation $\phi_{enc}(\cdot)$ to be comparatively low-dimensional, $\phi_{enc}(\mathbf{I}_t)$ or $(\phi_{enc}(\mathbf{I}_t), \mathbf{v}_t)$ could serve as a simplified but sufficient representation of the state, facilitating accelerated learning. Figure~\ref{ae_architecture} shows a schematic illustration of the architectures. 



We implemented and evaluated two different AEs. The first one is a regular convolutional AE. It uses the following encoder and decoder:
\begin{align}
\phi^{conv}_{enc} \equiv D_1 \circ C_3 \circ C_2 \circ C_1 \\
\phi^{conv}_{dec} \equiv C_6 \circ C_5 \circ C_4 \circ D_2
\end{align}
where the $C_i$ are convolutional layers and the $D_i$ are fully connected layers.

The second AE is based on the deep SAE described in~\cite{finn2015deep}, but with some significant variations. In the following sections, we refer it as Spatial Auto-encoder Variant (SAEV). The SAEV uses the encoder
\begin{align}
\phi_{enc}^{saev} \equiv S \circ C_3 \circ C_2 \circ C_1
\end{align}
where $C_i$ are convolutional layers. $C_1$,$C_2$ using exponential linear units ($ELU$) activation~\cite{clevert2015fast}, while $C_3$ uses a spatial softmax-activation:

\begin{align}
softmax(z)_{i,j,c} = \frac{e^{z_{i,j,c}}}{ \sum_{w=0}^W \sum_{h=0}^H e^{z_{w,h,c}} }
\end{align}

The mapping $S$ takes a number of feature maps, which it treats as bivariate probability distributions. For each, a feature location is estimated by the expectation values:
\begin{align}
x_c &= \mathbb{E}_{(i,j)\sim P_c} \left [ ~ i ~ \right ] \nonumber \\
y_c &= \mathbb{E}_{(i,j)\sim P_c} \left [ ~ j ~ \right ]
\end{align}

where $P_c(i,j)$ is the $(i,j)$ coordinate of the $c$\textsuperscript{th} feature-map of the input. The \emph{presence} of a feature is defined as the weighted sum
\begin{equation}
\rho_c = \sum_{i=0}^W \sum_{j=0}^H P_c(i,j) \cdot \mathcal{N}(i,j | \boldsymbol{\mu}=(x_c, y_c), \mathbf{\Sigma}= k \cdot \mathbf{I} )
\end{equation}
Intuitively, a feature map which is highly localized around the estimated position has a presence near $1$, whereas one that is very spread out will have presence close to $0$. The output from $S$ is the concatenation of the $(x_c, y_c, \rho_c)$ of each feature map. In other words, the intermediate representation contains actual image-coordinates of the features. 

The main difference between our SAEV architecture and the SAE described in~\cite{finn2015deep} is the decoder. The decoder we use is
\begin{align}
\phi_{dec}^{saev}\equiv B \circ C_6 \circ C_5 \circ C_4 \circ \Delta
\end{align}
where $C_i$ are convolutional layers, $C_4$, $C_5$ uses ELU-activations, while $C_6$ uses a sigmoid activation. $\Delta : \mathbb{R}^{N\times 3} \rightarrow \mathbb{R}^{W\times H \times N}$ is a transformation that takes the $N$ $(x_c,y_c,\rho_c)$-tuples and maps each to a feature map:
\begin{align}
\Delta( x_1,\ldots,x_C, y_1,\ldots,y_C, \rho_1,\ldots,\rho_C, )_{i,j,c}&= \notag\\
ELU(\rho_c - \Vert (i,j)-(x_c,y_c)\Vert_2)&
\end{align}

This creates $N$ feature maps, with peaks at $(i,j)=(x_c, y_c)$ that decrease radially outwards according to the ELU ~\cite{clevert2015fast}. To the output of $\Delta$, three convolutional layers are applied, followed by an addition operation with a trainable constant to complete the decoder. The constant addition operation frees up the prior stages of the architecture to focus on learning positions of things that are not always in the same place. 

\begin{figure}[h]
\centering
\includegraphics[width=8.6cm]{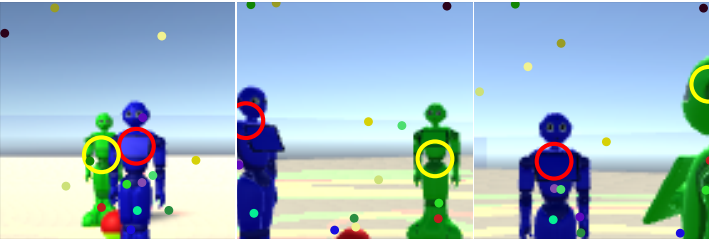}
\caption{Positions extracted from $\phi_{enc}(\mathbf{s}_t)$ visualized on top of their corresponding input images. The coloring of a feature is consistent across the three images. Most features are visualized as dots, except two features that have been chosen to be visualized as circles to more clearly show how they track their intended objects.} \label{visual_features}
\end{figure}

All models are trained using the Adam-optimizer ~\cite{kingma2014adam} on a loss function consisting of three components: reconstruction error $L_{err}=\Vert \phi(\mathbf{s}_t) - \mathbf{s}_t \Vert_2$, a presence based loss $L_{pre}=1-\rho(\mathbf{s}_t)$ that encourages localized features, and the smoothness loss $L_{smooth} = (\phi_{enc}(\mathbf{s_{t+1}}) - \phi_{enc}(\mathbf{s_{t}})) - (\phi_{enc}(\mathbf{s_{t}}) - \phi_{enc}(\mathbf{s_{t-1}}))$ defined in~\cite{finn2015deep}. For the convolutional AE, the presence loss is ill-defined and thus that term was omitted. One can now use the intermediate representation $\phi_{enc}(\mathbf{s}_t)$ as input to the RL framework, or to visualize the corresponding image coordinates, as is shown in Figure~\ref{visual_features}.

\subsection{Modeling Group Behavior}\label{subsection:approaching}

In a realistic multi-party conversation group, the individuals within it stand in appropriate positions with respect to others. This positional and orientational relationship has been defined as an \textit{F-formation} as proposed by Kendon~\cite{kendon1990conducting}. It characterizes a group of two or more individuals, typically in a conversation, to share information and interact with each other. Most importantly, it defines the \textit{o-space} which is a common focused space in the group in which all individuals look inward and is exclusive to those external. When conditions change, such as a new individual joining the group, the group members should change position or orientation in order to form a new group including the newcomer. Jan et al.~\cite{jan2007dynamic} proposed a group model which simulates these behaviors by a social force field. In this paper, we use an extended SSFM which maintains F-formation through repositioning and reorientating by a conversation force field. This force field is produced and updated by three forces: a repulsion force, an equality force, and a cohesion force. The details of social force fields are described in \cite{pedica2008social}. In order to better model conversation groups, Hall's proxemics theory \cite{hall1968proxemics} is adopted when generating social force fields, i.e. the repulsion, equality and cohesion forces occurring in personal, social and public spaces, respectively. 

The repulsion force prevents other agents from stepping inside its personal space and generates a repulsion force to push them away. Let $N_p$ be the number of other agents inside the personal space of agent $i$, and $\mathbf{p}_i$ is the corresponding position of agent $i$. The repulsion force is shown in equation~\ref{eq:repulsion}.

\begin{equation}\label{eq:repulsion}
\mathbf{F}_r = -(d_{p} - d_{min})^2\frac{\mathbf{p}_r}{||\mathbf{p}_r||}
\end{equation}

where $\mathbf{p}_r=\sum^{N_p}_i(\mathbf{p}_i-\mathbf{p})$, $\mathbf{p}$ is the position of the agent currently being evaluated. $d_{p}$ is the radius of its personal space, and $d_{min}$ is the distance to its closet agent inside the personal space. 

The equality force keeps o-space shared to all group members by generating an attraction or a repulsion force towards a point in o-space. Also, an orientation force towards o-space is generated to change body orientation. Let $N_s$ be the number of other agents inside the social space. The equality force $\mathbf{F}_e$ and equality orientation $\mathbf{d}_e$ are shown in equation~\ref{eq:equality}. 

\begin{align}\label{eq:equality}
\begin{split}
\mathbf{F}_e  &= (1-\frac{m}{||\mathbf{c}-\mathbf{p}||})(\mathbf{c}-\mathbf{p})
\\
\mathbf{d}_{e} &= \sum_{i}^{N_s}(\mathbf{p}_i-\mathbf{p})
\end{split}
\end{align}

where $c$ is the centroid, i.e. $\mathbf{c}=(\mathbf{p}+\sum^{N_s}_i\mathbf{p}_i)/(N_s+1)$, and $m$ is the mean distance of the members from the centroid. 

The cohesion force prevents an agent to be isolated from a group and keeps agents close to each other by generating an attraction force. Let $N_a$ be the number of other agents inside the public area, $o$ is the conversation center and $s$ is the radius of the o-space. The cohesion force $\mathbf{F}_{c}$ and cohesion orientation $\mathbf{d}_{c}$ are shown in equation~\ref{eq:cohesion}. 

\begin{align}\label{eq:cohesion}
\begin{split}
\mathbf{F}_{c}  &= \alpha (1-\frac{s}{||\mathbf{o}-\mathbf{p}||})(\mathbf{o}-\mathbf{p})
\\
\mathbf{d}_{c} &= \sum_{i}^{N_a}(\mathbf{p}_i-\mathbf{p})
\end{split}
\end{align}

where $\alpha = N_a/(N_s+1)$, which is the scaling factor for the cohesion force used to reduce the magnitude of the cohesion force if the agent is surrounded by other agents in its social area.

In order to include a component in reward function to drive the robot to approach the group. We incorporate the extended SSFM described previously and consider a line integral $r_1$ over a path $L$ in aforementioned force fields, namely force fields in personal, social and public spaces, to be the \textit{group forming reward}. Mathematically, the group forming reward for the robot agent is defined as follows

\begin{align}
R_1 &= \int_L r_1(\mathbf{u})\cdot d\mathbf{u}
\end{align}


where $\mathbf{u}$ is the position of the robot along the trajectory $L$, and $r_1(\mathbf{u}) = \sum_{i\in\{r,e,c\}}\mathbf{F}_{i}(\mathbf{u}) $ is the combined force on the robot agent. 
Note that the force fields $\mathbf{F}_{i}$ depend on the positioning of all agents, including the SHAs, but for notational simplicity, this is not made explicit in the formulae.

Together with the group forming reward, another reward function called \textit{non-increasing reward} is added to ensure the the energy in the force field is non-increasing. Mathematically, it is defined as 

\begin{align}
R_2 &= \int_{t_0}^{t_1} \mathds{1}_\mathcal{A}(\mathbf{u}(t)) dt 
\end{align}
where $\mathds{1}$ is the indicator function and $\mathcal{A}$ is the set of points along the robot's trajectory where ${dr_1(\mathbf{u}(t))}/{dt} \ge 0$.
These two reward functions help the robot agent to approach the group center. To add further incentive to complete the task, two other other reward components are added. They are a time-penalty
$ R_3 = -\int_{t_0}^{t_1}dt$ ($t_0$, $t_1$ are the times an episode starts and ends), together with a bonus reward $R_4$ for successful approaching behavior within the required number of time steps.

\subsection{Following Social Norms} \label{subsection:socialnorms}
In order to make the robot adhere to social norms when it is approaching the group, simulated feedback from other agents is taken into consideration. Therefore, the robot agent considers the impact of its own behavior on others, which is important in generating appropriate real-world robot approaching behaviors. Here, we define summation of all the line integrals of SHAs' paths in the force fields, 
\begin{align}
R_5 = -\sum_{j=0}^{N_p}\int_{L_j}\sum_{i\in\{r,e,c\}}\mathbf{F}_{ij} \boldsymbol{\cdot} d\mathbf{u}_j
\end{align}
where $N_p$ means the total number of the SHAs.
%


The final reward is a combination of all five rewards. Each is associated with a weight $w_i \ge 0$ to indicate the importance of that reward category. On top of the weights considered for each category of rewards, two other weights are used to influence the behavior of the robot. One weight is called \textit{egoism wight} $w_e$,  which decides how much the robot agent considers achieving its own goal of approaching the group center. The other weight, \textit{altruism weight} $w_a$ decides how much it cares about other agents, meaning avoiding pushing other SHAs around. The final reward is defined as follows: 
\begin{align}
R = & w_e \cdot (w_1 \cdot R_1 + w_2 \cdot R_2 + w_3\cdot R_3 + w_4\cdot R_4) + \notag\\
    &+ w_a \cdot w_5 \cdot R_5
\end{align}


By balancing the different weights, we produce a realistic reward function that captures important notions from human social interaction, such as respecting the private space of others.

\begin{figure}[h]
\centering
\includegraphics[align=c,width=1\linewidth]{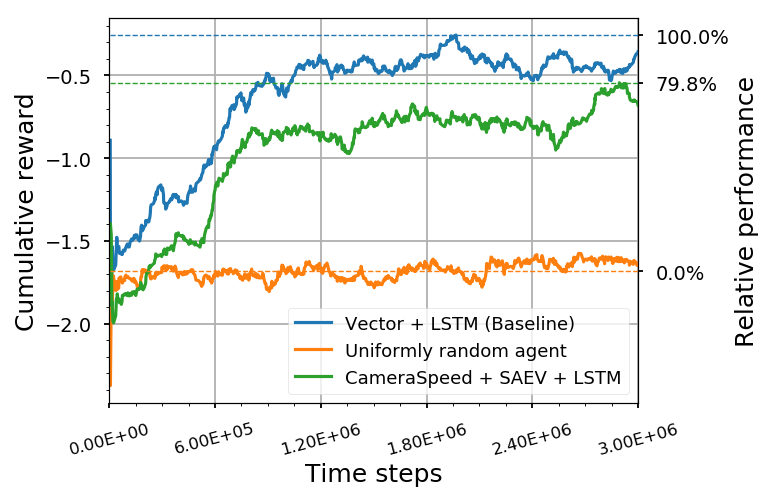}
\caption{Learning curves of the best model compared to a uniformly random agent and the baseline.} \label{learning_curve}
\end{figure}

\section{Results}
We used a DRL algorithm called Proximal Policy Optimization (PPO)~\cite{schulman2017proximal} to learn an appropriate behavior for the robot agent. We selected PPO due to its stability advantages~\cite{henderson2017deep} over DQN-based RL algorithms. We used ML-Agents Toolkit\footnote{https://github.com/Unity-Technologies/ml-agents}~\cite{juliani2018unity} to carry out our experiments.

\subsection{Models Configurations}
To determine what state representation and type of network structure for the value and policy networks are the most suitable for robot approaching behavior, we evaluate combinations of state representations, and network architectures. For the state representations containing visual information, we evaluate both AEs (\emph{conv} and \emph{SAEV} from section~\ref{section_staterepresentation}). The network structures considered are Feed-Forward (FF) networks and LSTM
networks. Table~\ref{results} shows the model configurations and their corresponding performance.

\begin{table}[ht]
\begin{center}
\begin{tabular}{||l r r ||}
\hline
\textbf{Model} & Reward & Percentage \\ [0.5ex]
\hline\hline
Vector + LSTM (Baseline)       & \textbf{-0.256} & \textbf{100.00}\% \\ \hline
CameraOnly + SAEV + FF         & -0.869 & 57.06\% \\ \hline
CameraOnly + SAEV + LSTM       & -0.804 & 61.63\% \\ \hline
CameraOnly + conv + FF         & -0.810 & 61.18\% \\ \hline
CameraOnly + conv + LSTM       & -1.091 & 41.51\% \\ \hline
CameraSpeed + SAEV + LSTM      & \textbf{-0.544} & \textbf{79.80}\% \\ \hline
CameraSpeed + conv + LSTM      & -0.709 & 68.22\% \\ \hline
Random policy                  & -1.684 & 0.00\% \\ \hline
\end{tabular}
\end{center}
\caption{Results of the different configurations. The reported results are the best cumulative reward of the model.}
\label{results}
\end{table}
Performance is measured both as cumulative reward (an exponentially weighted running average is used to smooth the function.), described in Section \ref{section_preliminaries}, and as percentages. Percentages express relative performance, such that $100\%$ correspond to the baseline performance, and $0\%$ to the mean performance of a uniformly random agent. Figure~\ref{learning_curve} shows the learning curve of the best model, which uses image and robot's speed as input, output of SAEV as learning state representation and a LSTM as policy network. 


\subsection{Approaching Behavior: Perceptual Study}\label{perceptual_study}
\begin{figure}[h] 
\centering
\includegraphics[align=c,width=.48\linewidth]{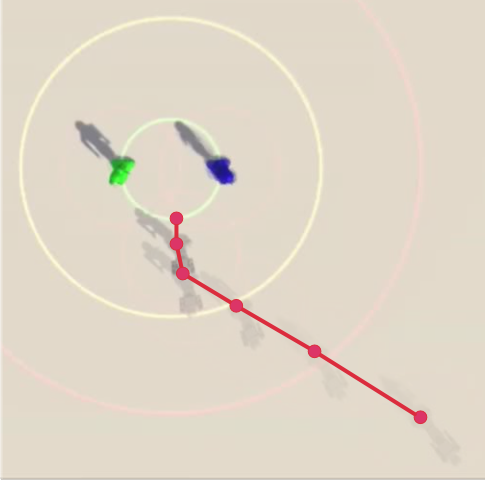}
\includegraphics[align=c,width=.48\linewidth]{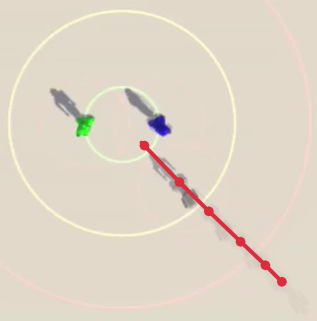}

\caption{Comparison of generated paths visualized as overlay screen shots of videos. The left image shows a sampled path generated by our model and the right image shows a sampled path generated by the SFFM with the same initial position.}\label{paths}
\end{figure}

We compare the robot approaching behavior learned by our model with the one generated by SFFM~\cite{pedica2008social}. In a study conducted by Pedica et al.~\cite{pedica2010avatars}, it was shown that SFFM increased believability of static group formation. A major drawback of SFFM is that it is directly controlled by the social forces and therefore does not act according to the current situation of the environment. We hypothesize that a learned robot agent that is able to accelerate and decelerate based on the simulated social feedbacks in RL framework can introduce more believability and social appropriateness. In order to compare the behaviors generated according to the SFFM with those generated with our proposed model, we implemented a version of SFFM and compared it with a model learned with the reward function defined in Section~\ref{subsection:socialnorms}. Figure~\ref{paths} shows paths sampled from our trained model and paths sampled from the SFFM with the same initial positions. One thing we  note here is that, though it is not the case in this study, a smoothing algorithm can also be applied to the learned policy to make the approaching behavior better.


In order to evaluate the behavior of our learned model compared to the behavior generated by SFFM, we conducted a perceptual study to evaluate the approaching behaviors using subjective measures. In this study we are interested in three dimensions of social appropriateness, namely \textit{polite}, \textit{sociable} and \textit{rude}, as in~\cite{okal2016learning}.
We created six videos of approaching behaviors in the simulated environment from a top-down view. The videos show six different approaching behaviors of the robot towards groups from three starting locations by both our model and the SFFM (Figure~\ref{paths}). Twenty participants (engineering students with a mixed cultural background; average age: 28.25 years) were asked to watch the videos and answer four questions for each video. Specifically, participants were asked to rate how much they thought the behaviors were \textit{polite}, \textit{sociable}, \textit{rude} and \textit{human-like}, using a  1-7  Likert  scale,  where  \textit{1} means "not at all" and \textit{7} means "very". The videos and their corresponding questions are given to the participants in a random order.

\begin{figure}[h]
\centering
\includegraphics[align=c,width=.70\linewidth]{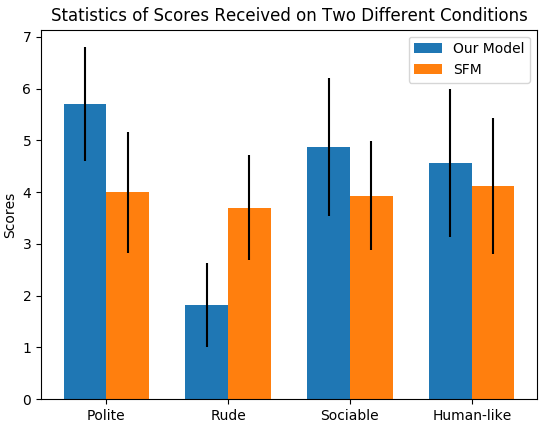}
\caption{Comparison of behaviors generated by the two models. The behavior generated by our model is generally considered to be more polite, less rude and more sociable.} \label{human_results}
\end{figure}

 Figure~\ref{human_results} shows participants' ratings of approaching behaviors generated by the two models. We found that people consider the behavior generated by our model to be significantly more polite ($t(19) = 6.45, p < .001$), less rude ($t(19) = 6.46, p < .001$) and more sociable ($t(19) = 2.65, p < .025$). However, we did not find the approaching behavior generated by our model to be significantly more human-like than the ones generated by SFFM ($t(19) = 1.01, p > .05$). This might be related to the fact that human-likeness is hard to measure when there are more than one factor involved, e.g. the agent's appearance ~\cite{macdorman2006subjective}, in addition to its movement.

\subsection{Approaching Behavior: Pilot User Study with Physical Robot}
\begin{figure}[h]
\centering
\includegraphics[align=c,width=.70\linewidth]{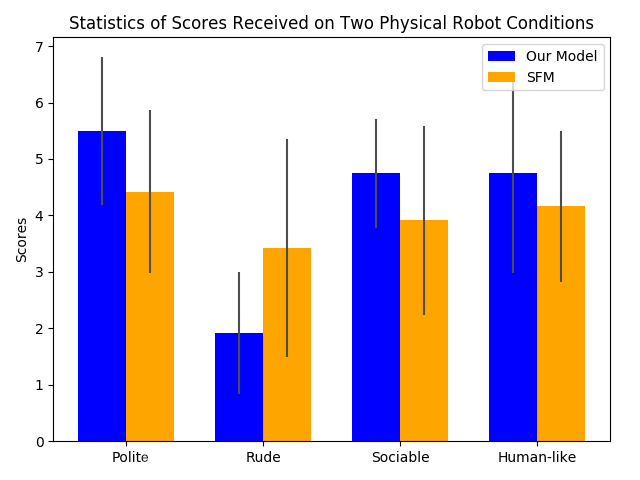}
\caption{Comparison of behaviors generated by the two models on a real robot. The behavior generated by our model is generally considered to be more polite, less rude and more sociable. } \label{physical_robot_results}
\end{figure}

We implemented robot approaching behaviors learnt with our model in a physical Pepper robot and conducted a user study with human participants to evaluate the model's performance in a real environment. In the study, a Pepper robot approaches a group of two people facing each other. Each group consists of a participant and an experimenter. We used the same questionnaire as in the Section~\ref{perceptual_study} to evaluate whether we obtain similar results to the perceptual study.
Twelve participants (mostly computer science students with mixed cultural background; average age: 31.1), were asked to evaluate two conditions in a within-subject design, namely robot approaching the group using the SFFM (condition one) and robot approaching the group according to our proposed model (condition two). For each condition, they were asked to first experience the robot's approaching behavior from one of two positions in the group (e.g., position A in Figure~\ref{physical_paths}) and then to switch position with the experimenter and experience the robot's approaching behavior from this position (e.g., position B in Figure~\ref{physical_paths}). During the study, participants interacted with the two conditions in a random order. After each condition, they were asked to fill in the questionnaire. We found that the robot's approaching behavior generated according to our model was perceived as significantly more polite ($t(11) = 2.399, p < .05$), less rude ($t(11) = 3.095, p < .05$) and more sociable ($t(11) = 2.278, p < .05$) than the one generated according to SFFM, but we did not find any significant difference for human-likeness ($t(11) = 0.7361, p > .05$). This is in line with the results from the perceptual study.  Figure~\ref{physical_robot_results} shows more detailed results.


\section{Discussion}
There are several things to be considered while using this approach to build a prior model. One of the main things is the necessity of using simulation. While it does need a well-established model like SFM to form a reward function using current RL technology, the generated behavior using RL is much richer. Also, when more advanced techniques are used, e.g. self-play or learning using a sparse reward, the agent may not need established models any more. One of the other questions could be is the state representation really needed? In this study, we specifically used an architecture similar to spatial AE~\cite{finn2015deep}. This architecture is easily transferable to the real world. Additionally, it is of importance to see, using these learned features, can we get similar results as using positions of the SHAs. Using a camera is a normal setup in real-world HRI scenarios.

\section{Conclusion and Future Work}
In this work, we proposed a deep learning scheme (SSBL) that can be considered as a general framework for social robot learning. As a demonstrator, we implemented a robot approaching behavior task based on this scheme. 
We designed a reward function combining concepts from SFFM and Hall's proxemics theory to enable the robot agent to learn a dynamical model which takes social norms into account. 
We found that SAEV outperforms the vanilla convolutional AE on this task with video input along or with video and speed information together given as input. Moreover, results from a perceptual study and a HRI study with a physical robot show that our model can generate more socially appropriate approaching behavior than SFFM. 

Future work will include a larger-scale study where human participants are asked to qualitatively assess the behavior of our learned model compared to the behavior generated by SFFM in real-world situations. Regarding the model configuration experiments, we will also investigate how to utilize more subtle real-world human feedback such as user engagement to refine our learned model using model-based RL algorithms. The expectation is that, by taking user affective and social behavior into account, robots will exhibit more socially appropriate approaching behavior. The next step in this process is to conduct policy refinement experiments through learning from subsequent real-world interaction with a physical robot interacting with humans.






\section*{Acknowledgement}
This work was supported by the COIN project (RIT15-0133) funded by the Swedish Foundation for Strategic Research and by the Swedish Research
Council (grant n. 2015-04378)

\balance
\bibliography{bluecoast} 
\bibliographystyle{ieeetr}

\end{document}